\documentclass[letterpaper, 10 pt, conference]{ieeeconf}  

\IEEEoverridecommandlockouts                              

\usepackage{amsmath}
\usepackage{array}
\usepackage[hyphens]{url}
\usepackage{hyperref}
\usepackage[all]{hypcap}   

\usepackage{soul}
\usepackage[pdftex]{graphicx} 
\usepackage{graphics}
\graphicspath{{pictures/giulias-pics/}}
\DeclareGraphicsExtensions{.pdf,.png}
\usepackage{amssymb}  
\usepackage{array}
\usepackage{eqparbox}
\usepackage{amssymb}
\usepackage{wasysym}
\usepackage{caption}
\usepackage{subcaption}

\usepackage{booktabs}
\usepackage{stmaryrd}
\usepackage{pifont}

\usepackage{bbm}

\usepackage{algpseudocode}
\usepackage{algorithm}
\usepackage{algorithmicx}
\usepackage{ulem} 

\usepackage{adjustbox}
\usepackage{multirow}
\usepackage{array}
\usepackage{booktabs}
\usepackage[table]{xcolor}

\definecolor{Gray}{gray}{0.85}

\usepackage{multirow}

\usepackage{array}

\newcolumntype{M}[1]{>{\centering\arraybackslash}m{#1}}
\newcolumntype{N}{@{}m{0pt}@{}}

\newcommand{\argmin}{\operatornamewithlimits{argmin}}
\newcommand{\argmax}{\operatornamewithlimits{argmax}}

\renewcommand{\emph}{\textit}

\title{\LARGE \bf
Incremental Robot Learning of New Objects with Fixed Update Time}

\author{Raffaello Camoriano$^{*\dagger\ddagger\diamond}$, Giulia Pasquale$^{*\dagger\ddagger\diamond}$, Carlo Ciliberto$^{\diamond}$,\\ Lorenzo Natale$^{\dagger}$, Lorenzo Rosasco$^{\diamond\ddagger}$ and Giorgio Metta$^{\dagger}$%
\thanks{$^{*}$Equal contribution}
\thanks{$^{\dagger}$iCub Facility, Istituto Italiano di Tecnologia (IIT), Genoa, Italy. Email: {\tt\small \{name.surname\}@iit.it}}
\thanks{$^{\diamond}$LCSL, IIT and MIT, Cambridge, USA.}%
\thanks{$^{\ddagger}$DIBRIS, Universit\`{a} degli Studi di Genova, Genoa, Italy.}%
}

\makeatletter
\newcommand\fs@spaceruled{\def\@fs@cfont{\bfseries}\let\@fs@capt\floatc@ruled
  \def\@fs@pre{\vspace{1\baselineskip}\hrule height.8pt depth0pt \kern2pt}%
  \def\@fs@post{\kern2pt\hrule\relax}%
  \def\@fs@mid{\kern2pt\hrule\kern2pt}%
  \let\@fs@iftopcapt\iftrue}
\makeatother

\begin{document}

\maketitle
\thispagestyle{empty}
\pagestyle{empty}

\begin{abstract}
We consider object recognition in the context of lifelong learning, where a robotic agent learns to discriminate between a growing number of object classes as it accumulates experience about the environment.
We propose an incremental variant of the Regularized Least Squares for Classification (RLSC) algorithm, and exploit its structure to seamlessly add new classes to the learned model.
The presented algorithm addresses the problem of having an unbalanced proportion of training examples per class, which occurs when new objects are presented to the system for the first time. 

We evaluate our algorithm on both a machine learning benchmark dataset and two challenging object recognition tasks in a robotic setting. Empirical evidence shows that our approach achieves comparable or higher classification performance than its batch counterpart when classes are unbalanced, while being significantly faster.

\end{abstract}

\section{Introduction}

In order for autonomous robots to operate in unstructured environments, several perceptual capabilities are required. Most of these skills cannot be hard-coded in the system beforehand, but need to be developed and learned over time as the agent explores and acquires novel experience. As a prototypical example of this setting, in this work we consider the task of visual object recognition in robotics: Images depicting different objects are received one frame at a time, and the system needs to incrementally update the internal model of known objects as new examples are gathered.

In the last few years, machine learning has achieved remarkable results in a variety of applications for robotics and computer vision \cite{Krizhevsky2012,simonyan2014,schwarz2015}. However, most of these methods have been developed for off-line (or ``batch'') settings, where the entire training set is available beforehand. The problem of updating a learned model online has been addressed in the literature \cite{french1999,Sayed:2008:AF:1370975,duchi2011,goodfellow2013}, but most algorithms proposed in this context do not take into account challenges that are characteristic of realistic lifelong learning applications. Specifically, in online classification settings, a major challenge is to cope with the situation in which a novel class is added to the model. Indeed, $1)$ most learning algorithms require the number of classes to be known beforehand and not grow indefinitely, and $2)$ the imbalance between the few examples of the new class (potentially just one) and the many examples of previously learned classes can lead to unexpected and undesired behaviors \cite{elkan2001}. More precisely, in this work we theoretically and empirically observe that the new and under-represented class is likely to be ignored by the learned model in favor of classes for which more training examples have already been observed, until a sufficient number of examples are provided also for such class.

Several methods have been proposed in the literature to deal with class imbalance in the batch setting by ``rebalancing'' the misclassification errors accordingly \cite{elkan2001,steinwart2008,he2009}. However, as we point out in this work, rebalancing cannot be applied to the online setting without re-training the entire model from scratch every time a new example is acquired. This would incur in computational learning times that increase at least linearly in the number of examples, which is clearly not feasible in scenarios in which training data grows indefinitely. 

In this work we propose a novel method that learns incrementally both with respect to the number of examples and classes, and accounts for potential class unbalance. Our algorithm builds on a recursive version of Regularized Least Squares for Classification (RLSC) \cite{rifkin2002everything,rifkin2003regularized} to achieve fixed incremental learning times when adding new examples to the model, while efficiently dealing with imbalance between classes. We evaluate our approach on a standard machine learning benchmark for classification and two challenging visual object recognition datasets for robotics. Our results highlight the clear advantages of our approach when classes are learned incrementally.


The paper is organized as follows: Sec.~\ref{sec:related_work} overviews related work on incremental learning and class imbalance. In Sec.~\ref{sec:classprob_unbal} we introduce the learning setting, discussing the impact of class imbalance and presenting two approaches that have been adopted in the literature to deal with this problem. Sec.~\ref{sec:background} reviews the recursive RLSC algorithm. 
In Sec.~\ref{sec:proposed_algorithm} we build on previous Sec.~\ref{sec:classprob_unbal} and~\ref{sec:background} to derive the approach proposed in this work, which extends recursive RLSC to allow for the addition of new classes with fixed update time, while dealing with class imbalance. In Sec.~\ref{sec:experiments} we report on the empirical evaluation of our method, concluding the paper in Sec.~\ref{sec:conclusions}.

\section{Related Work}
\label{sec:related_work}

{\bf Incremental Learning.} The problem of learning from a continuous stream of data has been addressed in the literature from multiple perspectives. The simplest strategy is to re-train the system on the updated training set, whenever a new example is received~\cite{xiao2014,jain2014}. The model from the previous iteration can be used as an initialization to learn the new predictor, reducing training time. These approaches require to store all the training data, and to retrain over all the points at each iteration. Their computational complexity increases at least linearly with the number of examples.

Incremental approaches that do not require to keep previous data in memory can be divided in {\it stochastic} and {\it recursive} methods. Stochastic techniques assume training data to be randomly sampled from an unknown distribution and offer asymptotic convergence guarantees to the ideal predictor~\cite{duchi2011}. However, it has been empirically observed that these methods do not perform well when seeing each training point only once, hence requiring to perform ``multiple passes'' over the data  \cite{hardt2015,lin2016}. This problem has been referred to as the ``catastrophic effect of forgetting'' \cite{french1999}, which occurs when training a stochastic model only on new examples while ignoring previous ones, and has recently attracted the attention of the Neural Networks literature \cite{srivastava2013,goodfellow2013}.

Recursive techniques are based, as the name suggests, on a recursive formulation of batch learning algorithms. Such formulation typically allows to compute the current model in closed form (or with few operations independent of the number of examples) as a combination of the previous model and the new observed example~\cite{Sayed:2008:AF:1370975,laskov2006}. As we discuss in more detail in Sec. \ref{sec:background}, the algorithm proposed in this work is based on a recursive method.

{\bf Learning with an Increasing Number of Classes.} Most classification algorithms have been developed for batch settings and therefore require the number of classes to be known a priori. However, this assumption is often broken in incremental settings, since new examples could belong to previously unknown classes. The problem of dealing with an increasing number of classes has been addressed in the contexts of transfer learning or {\it learning to learn}~\cite{thrun1996}. These settings consider a scenario where $T$ linear predictors have been learned to model $T$ classes. Then, when a new class is observed, the associated predictor is learned with the requirement of being ``close'' to a linear combination of the previous ones~\cite{tommasi2010,tommasi2012,kuzborskij2013}. Other approaches have been recently proposed where a class hierarchy is built incrementally as new classes are observed, allowing to create a taxonomy and exploit possible similarities among different classes \cite{xiao2014,sunneol2016}. However, all these methods are not incremental in the number of examples and require to retrain the system every time a new point is received.

{\bf Class Imbalance.} The problems related to class imbalance were previously studied in the literature~\cite{elkan2001,he2009,steinwart2008} and are addressed in Sec.~\ref{sec:classprob_unbal}. Methods to tackle this issue have been proposed, typically re-weighting the misclassification loss~\cite{tommasi2010} to account for class imbalance. However, as we discuss in Sec.~\ref{sec:incremental_recoding} for the case of the square loss, these methods cannot be implemented incrementally. This is problematic, since imbalance among multiple classes often arises in online settings, even if temporarily, for instance when examples of a new class are observed for the first time.

\section{Classification Setting and the Effect of Class Imbalance}
\label{sec:classprob_unbal}


In this section, we introduce the learning framework adopted in this work and describe the disrupting effect of imbalance among class labels. For simplicity, in the following we consider a binary classification setting, postponing the extension to multiclass classification to the end of the section. We refer the reader to \cite{steinwart2008} for more details about the Statistical Learning Theory for classification. 

\subsection{Optimal Bayes Classifier and its Least Squares Surrogate}

Let us consider a binary classification problem where input-output examples are sampled randomly according to a distribution $\rho$ over $\mathcal{X}\times\{-1,1\}$. The goal is to learn a function $b^*:\mathcal{X}\to\{-1,1\}$ minimizing the overall {\it expected} classification error
\begin{equation}\label{eq:expected_classification_error}
b^* = \argmin_{b:\mathcal{X}\to\{-1,1\}} \ \ \ \int_{\mathcal{X}\times\{-1,1\}} \mathbf{1}(b(x) - y) ~ d\rho(x,y),
\end{equation}
given a finite set of observations $\{x_i,y_i\}_{i=1}^n$, $x_i\in\mathcal{X}$, $y_i\in\{-1,1\}$ randomly sampled from $\rho$. Here $\mathbf{1}(s) :\mathbb{R} \to \{0,1\} $ denotes the binary function taking value $0$ if $s=0$ and $1$ otherwise. The solution to Eq.~\eqref{eq:expected_classification_error} is called the {\it optimal Bayes classifier} and it can be shown to satisfy the equation
\begin{equation}\label{eq:optimal_bayes}
b^*(x) = \left\{\begin{array}{cc} 1 & \mbox{if} \ \rho(1|x) > \rho(-1|x) \\ 
-1 & \mbox{otherwise}
\end{array}\right.,
\end{equation}
for all $x\in\mathcal{X}$. Here we have denoted by $\rho(y|x)$ the {\it conditional} distribution of $y$ given $x$ and in this work we will denote by $\rho(x)$ the {\it marginal} distribution of $x$, such that by Bayes' rule $\rho(x,y) = \rho(y|x)\rho(x)$.
Computing good estimates of $\rho(y|x)$ typically requires large training datasets and is often unfeasible in practice. Therefore, a so-called {\it surrogate} problem (see \cite{steinwart2008,bartlett2006}) is usually adopted to simplify the optimization problem at Eq.~\eqref{eq:expected_classification_error} and asymptotically recover the optimal Bayes classifier. In this sense, one well-known surrogate approach is to consider the least squares expected risk minimization  
\begin{equation}\label{eq:expected_risk_minimization}
	f^* = \argmin_{f:\mathcal{X}\to\mathbb{R}} \ \ \ \int_{\mathcal{X}\times\{-1,1\}} (y  - f(x))^2 ~ d\rho(x,y).
\end{equation}
The solution to Eq.~\eqref{eq:expected_risk_minimization} allows to recover the optimal Bayes classifier. Indeed, for any $f:\mathcal{X}\to\mathbb{R}$ we have
\begin{align*}
	\int (y - f(x))^2d \rho(x,y) = \int \int (y - f(x))^2 d\rho(y|x) d\rho(x)\\
    = \int \left[ (1-f(x))^2\rho(1|x) + (f(x)+1)^2\rho(-1|x) \right] d\rho(x),
\end{align*}
which implies that the minimizer of Eq.~\eqref{eq:expected_risk_minimization} satisfies
\begin{equation}\label{eq:least-squares_solution}
    f^*(x) = 2\rho(1|x) - 1 = \rho(1|x) - \rho(-1|x)
\end{equation}
for all $x\in\mathcal{X}$. The optimal Bayes classifier can be recovered from $f^*$ by taking its sign: $b^*(x) = sign(f^*(x))$. Indeed, $f^*(x)>0$ if and only if $\rho(1|x)>\rho(-1|x)$.\\

\noindent{\bf Empirical Setting}. When solving the problem in practice, we are provided with a finite set $\{x_i,y_i\}_{i=1}^n$ of training examples. In these settings the typical approach is to find an estimator $\hat{f}$ of $f^*$ by minimizing the {\it regularized empirical risk}
\begin{equation}\label{eq:erm}
	\hat{f} = \argmin_{f:\mathcal{X}\to\mathbb{R}} \ \ \ \frac{1}{n} \sum_{i=1}^n (y_i - f(x_i))^2 + R(f),
\end{equation}
where $R$ is a so-called {\it regularizer} preventing the solution $\hat{f}$ to overfit. Indeed, it can be shown~\cite{steinwart2008,shawe2004} that, under mild assumptions on the distribution $\rho$, it is possible for $\hat{f}$ to converge in probability to the ideal $f^*$ as the number of training points grows indefinitely. In Sec.~\ref{sec:background} we review a method to compute $\hat{f}$ in practice, both in the batch and in the online settings.

\subsection{The Effect of Unbalanced Data}

The classification rule at Eq.~\eqref{eq:optimal_bayes} associates every $x\in\mathcal{X}$ to the class $y$ with highest likelihood $\rho(y|x)$. However, in settings where the two classes are not balanced this approach could lead to unexpected and undesired behaviors. To see this, let us denote $\gamma = \rho(y=1) = \int_\mathcal{X} d\rho(y=1,x)$ and notice that, by Eq.~\eqref{eq:optimal_bayes} and the Bayes' rule, an example $x$ is labeled $y=1$ whenever
\begin{equation}\label{eq:unbalanced_rule}
	\rho(x|1)>\rho(x|-1)\frac{(1-\gamma)}{\gamma}.
\end{equation}
Hence, when $\gamma$ is close to one of its extremal values $0$ or $1$ (i.e. $\rho(y=1) \gg \rho(y=-1)$ or vice-versa), one class becomes clearly preferred with respect to the other and is almost always selected.

In Fig.~\ref{fig:example_rebalance} we report an example of the effect of unbalanced data by showing how the decision boundary (white dashed curve) of the optimal Bayes classifier from Eq.~\eqref{eq:optimal_bayes} varies as $\gamma$ takes values from $0.5$ (balanced case) to $0.9$ (very unbalanced case).
As it can be noticed, while the classes maintain the same shape, the decision boundary is remarkably affected by the value of $\gamma$.

Clearly, in an online robotics setting this effect could be critically suboptimal for two reasons: $1$) We would like the robot to recognize with high accuracy even objects that are less common to be seen. $2$) In incremental settings, whenever a novel object is observed for the first time, only few training examples are available (in the extreme case, just one) and we need a loss weighting fairly also underrepresented classes.

\subsection{Rebalancing the Loss}

In this paper, we consider a general approach to ``rebalancing'' the classification loss of the standard learning problem of Eq.~\eqref{eq:expected_classification_error}, similar to the ones in~\cite{elkan2001,steinwart2008}. We begin by noticing that in the balanced setting, namely for $\gamma = 0.5$, the classification rule at Eq.~\eqref{eq:unbalanced_rule} is equivalent to assigning class $1$ whenever $\rho(x|1) > \rho(x|-1)$ and vice-versa. Here we want to slightly modify the misclassification loss in Eq.~\eqref{eq:expected_classification_error} to recover this same rule also in unbalanced settings. To do so, we propose to apply a weight $w(y)\in\mathbb{R}$ to the loss $\mathbf{1}(b(x) - y)$, obtaining the problem
\begin{equation*}\label{eq:weighted_expected_classification_error}
b^*_w = \argmin_{b:\mathcal{X}\to\{-1,1\}} \ \ \ \int_{\mathcal{X}\times\{-1,1\}} w(y)\mathbf{1}(b(x) -  y) ~ d\rho(x,y) .
\end{equation*}
Analogously to the non-weighted case, the solution to this problem is 
\begin{equation}\label{eq:optimal_bayes_weighted_pre}
b^*_w(x) = \left\{\begin{array}{cc} 1 & \mbox{if} \ \rho(1|x)w(1) > \rho(-1|x)w(-1) \\ 
-1 & \mbox{otherwise}
\end{array}\right . .
\end{equation}
In this work we take the weights $w$ to be $w(1) = 1/\gamma$ and $w(-1) = \frac 1 {1-\gamma}$. Indeed, from the fact that $\rho(y|x) = \rho(x|y)(\rho(y)/\rho(x))$ we have that the rule at Eq.~\eqref{eq:optimal_bayes_weighted_pre} is equivalent to
\begin{equation}\label{eq:optimal_bayes_weighted}
b^*_w(x) = \left\{\begin{array}{cc} 1 & \mbox{if} \ \rho(x|1) > \rho(x|-1) \\ 
-1 & \mbox{otherwise}
\end{array}\right . , 
\end{equation}
which corresponds to the (unbalanced) optimal Bayes classifier in the case $\gamma = 0.5$, as desired.

\begin{figure}[t]
\vspace{2mm}
\centering
\includegraphics[width=0.32\linewidth]{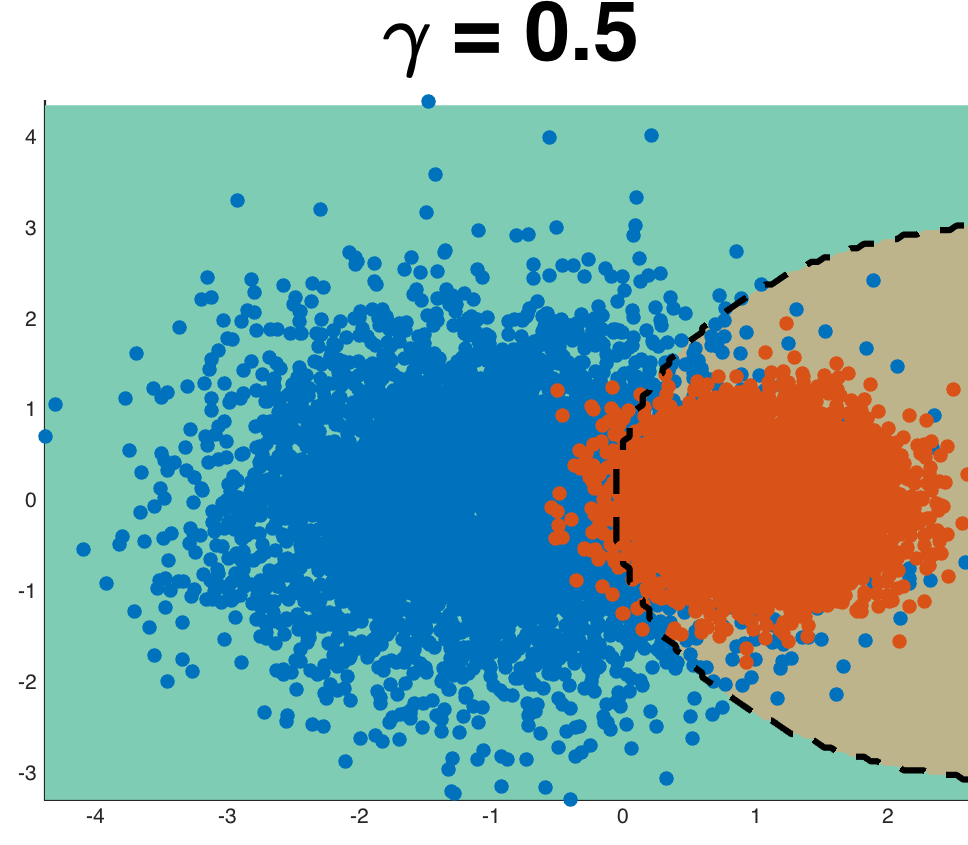}
\includegraphics[width=0.32\linewidth]{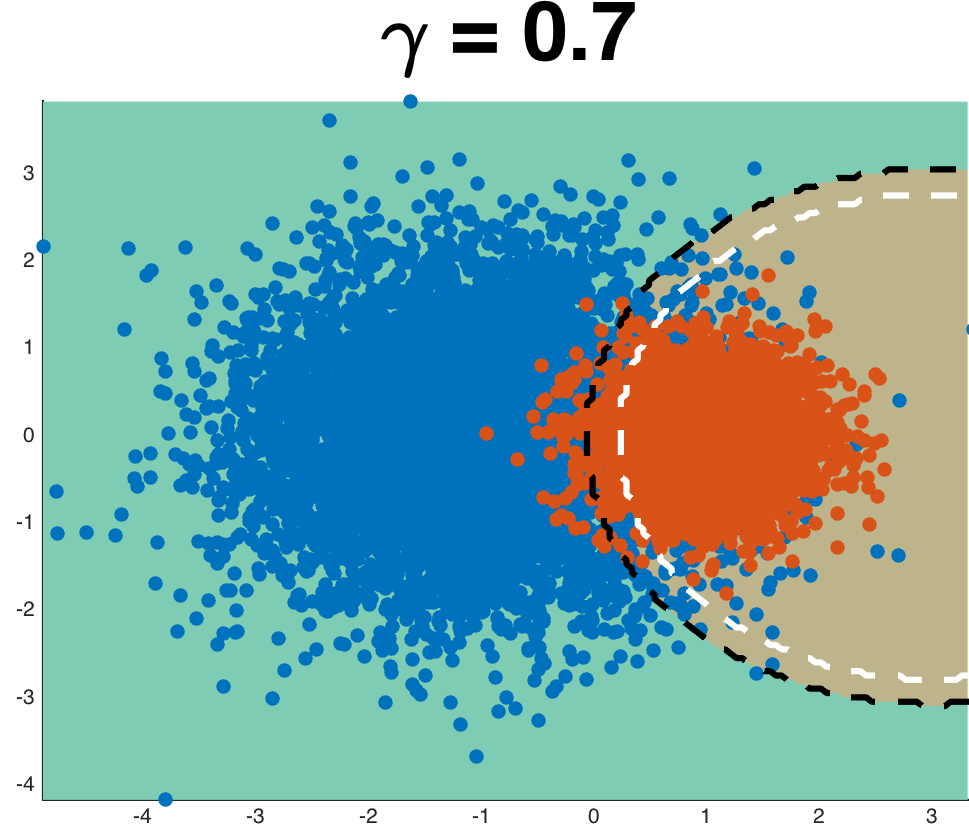}
\includegraphics[width=0.32\linewidth]{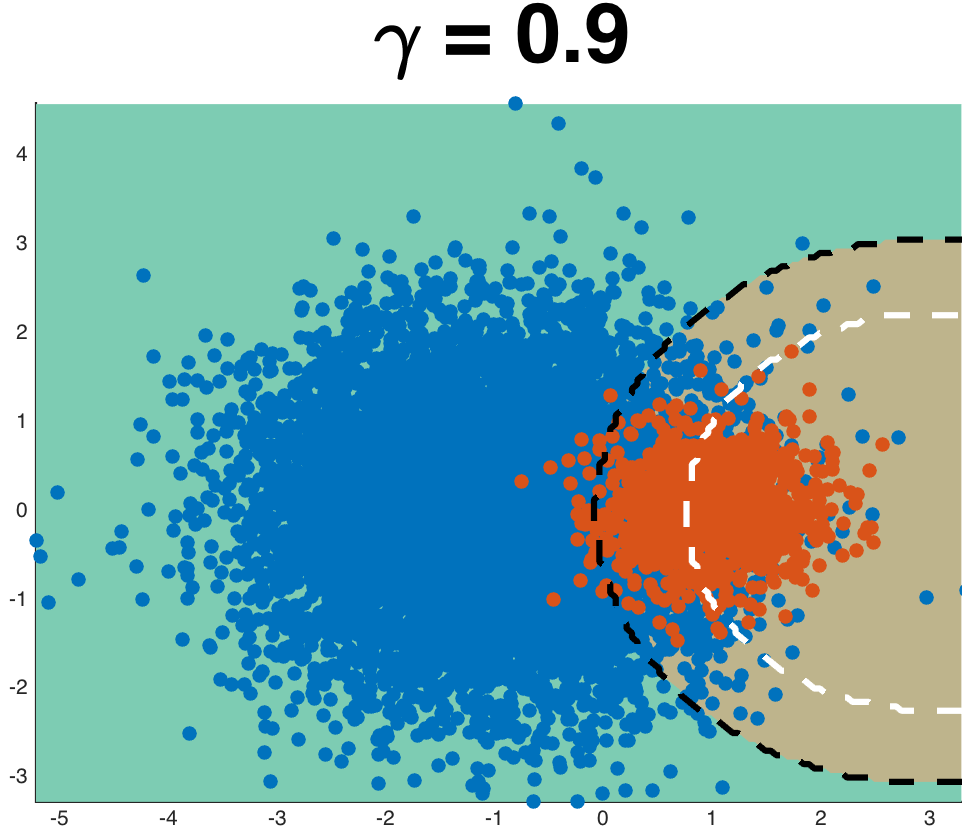}
\caption{Bayes decision boundaries for standard (dashed white line) and rebalanced (dashed black line) binary classification loss for multiple values of $\gamma=\rho(y=1)$ from $0.5$ to $0.9$. Data are sampled according to a Gaussian $\rho(x|y)\sim\mathcal{N}(\mu_y,\sigma_y)$ with $\mu_{1} = (-1,0)^\top$, $\mu_{-1} = (1,0)^\top$, $\sigma_{1}=1$ and $\sigma_{-1}=0.3$. The boundaries coincide when $\gamma=0.5$ (balanced data), while they separate as $\gamma$ increases.}
\label{fig:example_rebalance}
\end{figure}

Fig.~\ref{fig:example_rebalance} compares the unbalanced and rebalanced optimal Bayes classifiers for different values of $\gamma$. Notice that rebalancing leads to solutions that are invariant to the value of $\gamma$ (compare the black decision boundary with the white one).

\subsection{Rebalancing and Recoding the Least Squares Loss}\label{sec:rebalancing_recoding_least_squares}

Interestingly, the strategy of changing the weight of the classification error loss can be naturally extended to the least squares surrogate. If we consider the weighted least squares problem, 
\begin{equation}\label{eq:weigthed_least_squares_expectation}
	f_w^* = \argmin_{f:\mathcal{X}\to\mathbb{R}} \ \ \ \int_{\mathcal{X}\times\{-1,1\}} w(y)(y  - f(x))^2 d\rho(x,y),
\end{equation}
we can again recover the (weighted) rule $b_w^*(x) = sign(f_w^*(x))$ like in the non-weighted setting. Indeed, by direct calculation it follows that Eq.~\eqref{eq:weigthed_least_squares_expectation} has solution 
\begin{equation}\label{eq:sol_weighted_rls}
f^w_*(x) = \frac{\rho(1|x)w(1) - \rho(-1|x)w(-1)}{\rho(1|x)w(1) + \rho(-1|x)w(-1)}.
\end{equation}
If we assume $w(1)>0$ and $w(-1)>0$ (as in this work), the denominator of Eq.~\eqref{eq:sol_weighted_rls} is always positive and therefore $sign(f_w^*(x))>0$ if and only if $\rho(1|x)w(1)>\rho(-1|x)w(-1)$, as desired.\\

\noindent{\bf Coding}. An alternative approach to recover the rebalanced optimal Bayes classifier via least squares surrogate is to apply a suitable coding function to the class labels $y= \{-1,1\}$, namely 
\begin{equation}
\label{eq:coded_least_squares_expectation_binary}
	f_c^* = \argmin_{f:\mathcal{X}\to\mathbb{R}} \ \ \ \int_{\mathcal{X}\times\{-1,1\}} (c(y)  - f(x))^2 d\rho(x,y),
\end{equation}
where $c:\{-1,1\}\to\mathbb{R}$ maps the labels $y$ into scalar codes $c(y)\in\mathbb{R}$. Analogously to the unbalanced (and uncoded) case, the solution to Eq.~\eqref{eq:coded_least_squares_expectation_binary} is
\begin{equation}\label{eq:coding_solution}
f_c^*(x) = c(1)\rho(1|x) - c(-1)\rho(-1|x) ,
\end{equation}
which, for $c(y) = w(y)$, corresponds to the numerator of Eq.~\eqref{eq:sol_weighted_rls}. Therefore, the optimal (rebalanced) Bayes classifier is recovered again by $b_w^*(x) = sign(f_c^*(x))$.\\


\subsection{Multiclass Rebalancing and Recoding}\label{sec:multiclass}

In the multiclass setting, the optimal Bayes decision rule corresponds to the function $b^*:\mathcal{X}\to\{1,\dots,T\}$, assigning a label $t \in \{1,\dots,T\}$ to $x\in\mathcal{X}$ when $\rho(t|x)>\rho(s|x)$ $\forall s \neq t$, with $t,s \in \{1,\dots,T\}$. Consequently, the rebalanced decision rule would assign class $t$, whenever $\rho(y=t|x)w(t)>\rho(y=s|x)w(s)$ $\forall s \neq t$, where the function $w:\{1,\dots,T\}\to\mathbb{R}$ assigns a weight to each class. Generalizing the binary case, in this work we set $w(t) = 1/\rho(y=t)$, where we denote $\rho(y=t) = \int_\mathcal{X} d\rho(t,x)$, for each $t \in \{1,\dots,T\}$.

In multiclass settings, the surrogate least squares classification approach is recovered by adopting a {\it 1-vs-all} strategy, formulated as the vector-valued problem
\begin{equation}
\label{eq:coded_least_squares_expectation}
f^* = \argmin_{f:\mathcal{X}\to\mathbb{R}^T} \ \ \ \int_{\mathcal{X}\times\{1,\dots,T\}} \|e_y  - f(x)\|^2 d\rho(x,y) ,
\end{equation}
where $e_t\in\mathbb{R}^T$ is a vector of the canonical basis $\{e_1,\dots,e_T\}$ of $\mathbb{R}^T$ (with the $t$-th coordinate equal to $1$ and the remaining $0$). Analogously to the derivation of Eq.~\eqref{eq:least-squares_solution}, it can be shown that the solution to this problem corresponds to $f^*(x) = (\rho(1|x),\dots,\rho(T|x))^\top\in\mathbb{R}^T$ for all $x\in\mathcal{X}$. Consequently, we recover the optimal Bayes classifier by
\begin{equation}\label{eq:multiclass_bayes_recover}
	b^*(x) = \argmax_{t=1,\dots,T} f(x)_t,
\end{equation}
where $f(x)_t$ denotes the $t$-th entry of the vector $f(x)\in\mathbb{R}^T$.

The extensions of recoding and rebalancing approaches to this setting follow analogously to the binary setting discussed in Sec.~\ref{sec:rebalancing_recoding_least_squares}.
In particular, the coding function $c:\{e_1,\dots,e_T\}\to\mathbb{R}$ consists in mapping a vector of the basis $e_t$ to $c(e_t) = e_t/\rho(y=t)$.\\

\noindent{\bf Note}. In previous sections we presented the analysis on the binary case by considering a $\{-1,1\}$ coding for class labels. This was done to offer a clear introduction to the classification problem, since we need to solve a single least squares problem to recover the optimal Bayes classifier. Alternatively, we could have followed the approach introduced in this section where classes have labels $y=\{1,2\}$ and adopt surrogate labels $e_1 = [1,0]^\top$ and $e_2 = [0,1]^\top$. This would have led to training two distinct classifiers and choosing the predicted class as the $argmax$ of their scores, according to Eq.~\eqref{eq:multiclass_bayes_recover}. The two approaches are clearly equivalent since the Bayes classifier corresponds respectively to the inequalities $\rho(1|x)>\rho(-1|x)$ or $\rho(1|x)>\rho(2|x)$.

\section{RLSC and Recursive Formulation}
\label{sec:background}

In this section we review the standard algorithm for Regularized Least Squares Classification (RLSC) and its recursive formulation used for incremental updates.

\subsection{Regularized Least Squares for Classification}
\label{sec:rls}

We address the problem of solving the empirical risk minimization introduced in Eq. \eqref{eq:erm} in the multiclass setting. Let $\{x_i,y_i\}_{i=1}^n$ be a finite training set, with inputs $x_i\in\mathcal{X}=\mathbb{R}^d$ and labels $y_i\in\{1,\dots,T\}$. In this work, we will assume a linear model for the classifier $\hat{f}$, namely $f(x) = W^\top x$, with $W$ a matrix in $\mathbb{R}^{d \times T}$. We can rewrite Eq.~\eqref{eq:erm} in matrix notation as 
\begin{equation}\label{eq:matrix-ls}
	\widehat{W} = \argmin_{W\in\mathcal{R}^{d \times T}} \|Y - XW\|_F^2 + \lambda \|W\|_F^2
\end{equation}
with $\lambda>0$ the {\it regularization parameter} and $X\in\mathbb{R}^{n \times d}$ and $Y\in\mathbb{R}^{n \times T}$ the matrices whose $i$-th rows correspond respectively to $x_i\in\mathbb{R}^d$ and $e_{y_i}\in\mathbb{R}^T$. We denote by $\|\cdot\|_F^2$ the squared Frobenius norm of a matrix (i.e. the sum of its squared entries).

The solution to Eq.~\eqref{eq:matrix-ls} is
\begin{equation}\label{eq:W_solution}
\widehat{W} = (X^\top X + \lambda I_{d})^{-1}X^\top Y \in \mathbb{R}^{d \times T} ,
\end{equation}
where $I_{d}$ is the $d \times d$ identity matrix (see for instance \cite{boyd2004}).\\

\noindent{\bf Prediction}. According to the rule introduced in Sec.~\ref{sec:multiclass}, a given $x\in\mathbb{R}^d$ is classified according to 
\begin{equation}
\hat{b}(x) = \argmax_{i=1,\dots,T} \ \ \hat{f}(x)_i = \argmax_{i=1,\dots,T} \ \ (\widehat{W}^{(i)})^\top x ,
\end{equation}
with $\widehat{W}^{(i)}\in\mathbb{R}^d$ denoting the $i$-th column of $\widehat{W}$.

\subsection{Recursive Formulation}
\label{sec:recupdate}

The closed form for the solution at Eq. \eqref{eq:W_solution} allows to derive a recursive formulation to incrementally update $\widehat{W}$ in fixed time as new training examples are observed~\cite{Sayed:2008:AF:1370975}. Consider a learning process where training data are provided to the system one at a time. At iteration $k$ we need to compute $W_{k} = (X_{k}^\top X_{k} + \lambda I_d)^{-1} X_{k}^\top Y_{k}$, where $X_{k} \in \mathbb{R}^{k \times d}$ and $Y_{k} \in \mathbb{R}^{k \times T}$ are the matrices whose rows correspond to the first $k$ training examples. The computational cost for evaluating $W_{k}$ according to Eq.~\eqref{eq:W_solution} is $O(kd^2)$ (for the matrix products) and $O(d^3)$ (for the inversion). This is undesirable in an online setting where $k$ can grow indefinitely. To this end, we now review how $W_k$ can be computed incrementally from $W_{k-1}$ in $O(Td^2)$.
To see this, first notice that, by construction,
\[
X_{k} = [ X_{k-1}^\top , x_{k} ]^\top \qquad Y_{k} = [ Y_{k-1}^\top , e_{y_{k}} ]^\top ,
\]
and therefore, if we denote $A_k = X_k^\top X_k + \lambda I_d$ and  $b_k = X_k^\top Y_k$, we obtain the recursive formulations
\begin{align}
A_{k} &= X_{k}^\top X_{k} + \lambda I_{d} \nonumber \\
&=X_{k-1}^\top X_{k-1} + x_{k}^\top x_{k} + \lambda I_{d} \nonumber \\
&=A_{k-1} + x_{k}^\top x_{k} + \lambda I_{d}
\end{align}
and
\begin{equation}
b_{k} = X_{k}^\top Y_{k} = X_{k-1}^\top Y_{k-1} + x_{k}e_{y_k}^\top = b_{k-1} + x_k e_{y_k}^\top.
\end{equation}

Computing $b_k$ from $b_{k-1}$ requires $O(d)$ operations (since $e_{y_k}$ has all zero entries but one). Computing $A_k$ from $A_{k-1}$ requires $O(d^2)$, while the inversion $A_k^{-1}$ requires $O(d^3)$. To reduce the cost of the (incremental) inversion, we recall that for a positive definite matrix $A_k$ for which its Cholesky decomposition $A_k = R_k^\top R_k$ is known (with $R^k\in\mathbb{R}^{d \times d}$ upper triangular), the inversion $A_k^{-1}$ can be computed in $O(d^2)$ \cite{Golub1996}. In principle, computing the Cholesky decomposition of $A_k$ still requires $O(d^3)$, but we can apply a rank-one update to the Cholesky decomposition at the previous step, namely $A_k = R_k^\top R_k = R_{k-1}^\top R_{k-1} + x_k x_k^\top = A_{k-1} + x_k x_k^\top$, which is known to require $O(d^2)$~\cite{bjoerck_least_squares96}. Several implementations are available for the Cholesky rank-one updates; in our experiments we used the {\sc MATLAB} routine {\sc cholupdate}.

Therefore, the update $W_k$ from $W_{k-1}$ can be computed in $O(Td^2)$, since the most expensive operation is the multiplication $A_k^{-1}b_k$. In particular, this computation is independent of the current number $k$ of training examples seen so far, making this algorithm suited for online settings.


\section{Incremental RLSC with Class Extension and Recoding}
\label{sec:proposed_algorithm}
In this Section, we present our approach to incremental multiclass classification where we account for the possibility to extend the number of classes incrementally and apply the recoding approach introduced in Sec.~\ref{sec:classprob_unbal}. The algorithm is reported in Alg.~\ref{alg:incremental_learning}.

\subsection{Class Extension}
We propose a modification of recursive RLSC, allowing to extend the number of classes in constant time with respect to the number of examples seen so far. Let $T_{k}$ denote the number of classes seen up to iteration $k$.  We have two possibilities:
\begin{enumerate}
\item The new example $(x_k, y_k)$ belongs to one of the known classes, i.e. $e_{y_k} \in \mathbb{R}^{T_{k-1}}$, with $T_k = T_{k-1}$.
\item $(x_k, y_k)$ belongs to a new class, implying that $y_k = T_k = T_{k-1} + 1$.
\end{enumerate}

In the first case, the update rules for $A_k$, $b_k$ and $W_k$ explained in Section \ref{sec:recupdate} can be directly applied. In the second case, the update rule for $A_k$ remains unchanged, while the update of $b_k$ needs to account for the increase in size (since $b_k\in\mathbb{R}^{k \times (T_{k-1} + 1)}$). However, we can modify the update rule for $b_k$ without increasing its computational cost by first adding a new column of zeros $\mathbf{0}\in\mathbb{R}^d$ to $b_{k-1}$, namely
\begin{equation}
b_k = [b_{k-1}, \mathbf{0}] + x_k^\top e_{y_k} ,
\end{equation}
which requires $O(d)$ operations. Therefore, with the strategy described above it is indeed possible to extend the classification capabilities of the incremental learner during online operation, without re-training it from scratch. In the following, we address the problem of dealing with class imbalance during incremental updates by performing incremental recoding.

\floatstyle{spaceruled}
\restylefloat{algorithm}
\begin{algorithm}[t]
\caption{Incremental RLSC with Class Recoding}
  \begin{algorithmic}
   \Statex \textbf{Input:} Hyperparameters $\lambda > 0$, $\alpha \in [0,1]$
    \Statex \textbf{Output:} Learned weights $W_k$ at each iteration
    \Statex \textbf{Initialize:} $R_0 \gets \sqrt{\lambda}I_d, ~ b_0 \gets \emptyset, ~ \gamma_0 \gets \emptyset, ~ T \gets 0$ \\
    
    \algblock[Increment]{Increment}{EndIncrement}
    
    \algnewcommand\algorithmicincrement{\textbf{Increment: }}
    \algrenewtext{Increment}{\algorithmicincrement}
    \algnewcommand\algorithmicendincrement{\textbf{end\ increment}}
    \algrenewtext{EndIncrement}{}

    \Increment Observe input $x_k\in\mathbb{R}^d$ and output label $y_k$:
        \If{  ($y_k  =  T + 1$) }
            \State $T \gets T + 1$
            \State $ \gamma_{k-1} \gets [ \gamma_{k-1}^\top, 0]^\top$
            \State $b_{k-1} \gets \left[b_{k-1}, \mathbf{0}\right]$, with $\mathbf{0}\in\mathbb{R}^d$
        \EndIf  
        \State $\gamma_k \gets \gamma_{k-1} + e_{y_k}$
        \State $\Gamma_k \gets k \cdot \text{diag}(\gamma_k)^{-1}$
        \State $b_k \gets b_{k-1} + x_k^\top e_{y_k}$
        \State $R_k \gets$ \textsc{choleskyUpdate}$(R_{k-1}, x_k)$
        \State $W_k \gets R_k^{-1} (R_k^\top)^{-1} b_k (\Gamma_k)^\alpha$
    \EndIncrement \Return $W_k$
  \end{algorithmic}
  \label{alg:incremental_learning}
\end{algorithm}

\subsection{Incremental Recoding}
\label{sec:incremental_recoding}

The main algorithmic difference between standard RLSC and the variant with recoding is in the matrix $Y$ containing output training examples. Indeed, according to the recoding strategy, the vector $e_{y_k}$ associated to an output label $y_k$ is coded into $c(e_{y_k}) = e_{y_k}/\rho(y=y_k)$. In the batch setting, this can be formulated in matrix notation as 
$$
W = (X^\top X + \lambda I_d)^{-1} X^\top Y \Gamma ,
$$
where the original output matrix is replaced by its encoded version $Y^{(c)}= Y\Gamma\in\mathbf{R}^{n \times T}$, with $\Gamma$ the $T \times T$ diagonal matrix whose $t$-th diagonal element is $\Gamma_{tt} = 1/\rho(y=t)$. Clearly, in practice the $\rho(y=t)$ are estimated empirically (e.g. by $\hat{\rho}(y=t) = n_t/n$, the ratio between the number $n_t$ of training examples belonging to class $t$ and the total number $n$ of examples).

The above formulation is favorable for the online setting. Indeed, we have 
\begin{equation}\label{eq:incremental_coding}
X_k^\top Y_k \Gamma_k = b_k \Gamma_k = (b_{k-1} + x_k^\top y_k) \Gamma_k,
\end{equation}
where $\Gamma_k$ is the diagonal matrix of the (inverse) class distribution estimators $\hat\rho$ up to iteration $k$. $\Gamma_k$ can be computed incrementally in $O(T)$ by keeping track of the number $k_t$ of examples belonging to $t$ and then computing $\hat\rho_k(y=t) = k_t/k$ (see Alg.\ref{alg:incremental_learning} for how this update was implemented in our experiments). 
Note that the above step requires $O(dT)$, since updating the (uncoded) $b_k$ from $b_{k-1}$ requires $O(d)$ and multiplying $b_k$ by a diagonal matrix requires $O(dT)$. All the above computations are dominated by the product $A_k^{-1}b_k$, which requires $O(Td^2)$. Therefore, our algorithm is computationally equivalent to the standard incremental RLSC approach.
\\

\noindent{\bf Coding as a Regularization Parameter.} Depending on the amount of training examples seen so far, the estimator $k_{t}/k$ could happen to not approximate $\rho(y=t)$ well. In order to mitigate this issue, we propose to introduce a parameter $\alpha\in[0,1]$ and raise $\Gamma_k$ element-wise to the power of $\alpha$ (indicated by $(\Gamma_k)^\alpha$). Indeed, it can be noticed that for $\alpha = 0$ we recover the (uncoded) standard RLSC, since $(\Gamma_k)^0 = I_T$, while $\alpha=1$ applies full recoding. In Sec.~\ref{sec:model_selection} we discuss an efficient heuristic to find $\alpha$ in practice.
\\

\noindent{\bf Incremental Rebalancing.} Note that the loss-rebalancing algorithm (Sec.~\ref{sec:rebalancing_recoding_least_squares}) cannot be implemented incrementally. Indeed, the solution of the rebalanced empirical RLSC is
\begin{equation}\label{sec:incr_rebalancing}
W_k = (X_k^\top \Sigma_k X_k + \lambda I_d)^{-1} X_k^\top \Sigma_K Y_k ,
\end{equation}
with $\Sigma_k$ a diagonal matrix whose $i$-th entry is equal to 
 $(\Sigma_k)_{ii} = 1/\hat{\rho}(y = t_i)$, with $t_i$ the class of the $i$-th training example. Since $\Sigma_k$ changes at every iteration, it is not possible to derive a rank-one update rule for $(X_k^\top \Sigma_k X_k + \lambda I_d)^{-1}$ as for the standard RLSC.

\section{Experiments}\label{sec:experiments}

We empirically assessed the performance of Alg.\ref{alg:incremental_learning} on a standard benchmark for machine learning and on two visual recognition tasks in robotics. To evaluate the improvement provided by the incremental recoding when classes are imbalanced, we compared the accuracy of the proposed method with the standard recursive RLSC presented in Sec.~\ref{sec:recupdate}. As a competitor in terms of accuracy, we also considered the rebalanced approach presented in Eq.~\eqref{sec:incr_rebalancing} (which, we recall, cannot be implemented incrementally).

\subsection{Experimental Protocol}
\label{sec:protocol}

We adopted the following experimental protocol\footnote{Code available at~\url{https://github.com/LCSL/incremental_multiclass_RLSC}}:
\begin{enumerate}
\item Given a dataset with $T$ classes, we simulated a scenario where a new class is observed by selecting $T-1$ of them to be ``balanced'' and the remaining one to be under-represented. 
\item We trained a classifier on the balanced classes, using a randomly sampled dataset containing $n_{bal}$ examples per class (specified below for each dataset). We sampled a validation set with $n_{bal}/5$ examples per class.
\item We incrementally trained the classifier from the previous step by sampling online $n_{imb}$ examples for the $T$-th class. Model selection was performed {\it using exclusively the validation set of the balanced classes}, following the strategy described in Sec. \ref{sec:model_selection}.
\item To measure performance, we sampled a separate test set containing $n_{test}$ examples per class (both balanced and under-represented) and we measured the accuracy of the algorithms on the test set while they were trained incrementally. 
\end{enumerate}

For each dataset, we averaged results over multiple independent trials randomly sampling the validation set. In Table~\ref{tab:accuracy} we report the test accuracy on the imbalanced class and on the entire test set.

\subsection{Datasets}\label{sec:datasets}

{\bf MNIST} \cite{lecun1998gradient} is a benchmark composed of $60$K $28 \times 28$ greyscale pictures of digits from $0$ to $9$. We addressed the 10-class digit recognition problem usually considered in the literature, but using $n_{bal}=1000$ training images per class. The test set was obtained by sampling $n_{test}=200$ images per class. 
We used the raw pixels of the images as inputs for the linear classifier.

{\bf iCubWorld28} \cite{pasquale2015teaching} is a dataset for visual object recognition in robotics, collected during a series of sessions where a human teacher showed different objects to the iCub humanoid robot \cite{Metta2010}. We addressed the task of discriminating between the $28$ objects instances in the dataset, using all available acquisition sessions per object and randomly sampling $n_{bal} = 700$ and $n_{test} = 700$ examples per class. 
We performed feature extraction as specified in \cite{pasquale2015teaching}, i.e. by taking the activations of the {\it fc$7$} layer of the {\sc CaffeNet} Convolutional Neural Network~\cite{Jia2014}.

{\bf RGB-D Washington} \cite{lai2011} is a visual object recognition dataset comprising $300$ objects belonging to $51$ categories, acquired by recording image sequences of each object while rotating on a turntable.  
We addressed the $51$-class object categorization task, averaging results over the ten splits specified in~\cite{lai2011} (where, for each category, a random instance is left out for testing). We subsampled one cropped RGB frame every five from the full dataset, following the standard procedure. We sampled $n_{bal}=500$ and $n_{test}=400$ images per class and performed feature extraction analogously to {\it iCubWorld28}, using the output of {\sc CaffeNet}'s $fc6$ layer.

\subsection{Model Selection}
\label{sec:model_selection}

In traditional batch learning settings for RLSC, model selection for the hyperparameter $\lambda$ is typically performed via hold-out, k-fold or similar cross-validation techniques. In the incremental setting these strategies cannot be directly applied since examples are observed online, but a simple approach to create a validation set is to hold out every $i$-th example without using it for training (e.g., we set $i=6$). At each iteration, multiple candidate models are trained incrementally, each for a different value of $\lambda$, and the one with highest validation accuracy is selected for prediction.

However, following the same argument of Sec.~\ref{sec:classprob_unbal}, in presence of class imbalance this strategy would often select classifiers that ignore the under-represented class. Rebalancing the validation loss (see Sec.~\ref{sec:classprob_unbal}) does not necessarily solve the issue, but could rather lead to overfitting the under-represented class, degrading the accuracy on other classes since errors count less on them. Motivated by the empirical evidence discussed below, in this work we have adopted a model selection heuristic for $\lambda$ and $\alpha$ in Alg.~\ref{alg:incremental_learning}, which guarantees to not degrade accuracy on well-represented classes, while at the same time achieving higher or equal accuracy on the under-represented one.

\begin{figure}[t]
\vspace*{2mm}
\centering
\includegraphics[width = 0.4\textwidth]{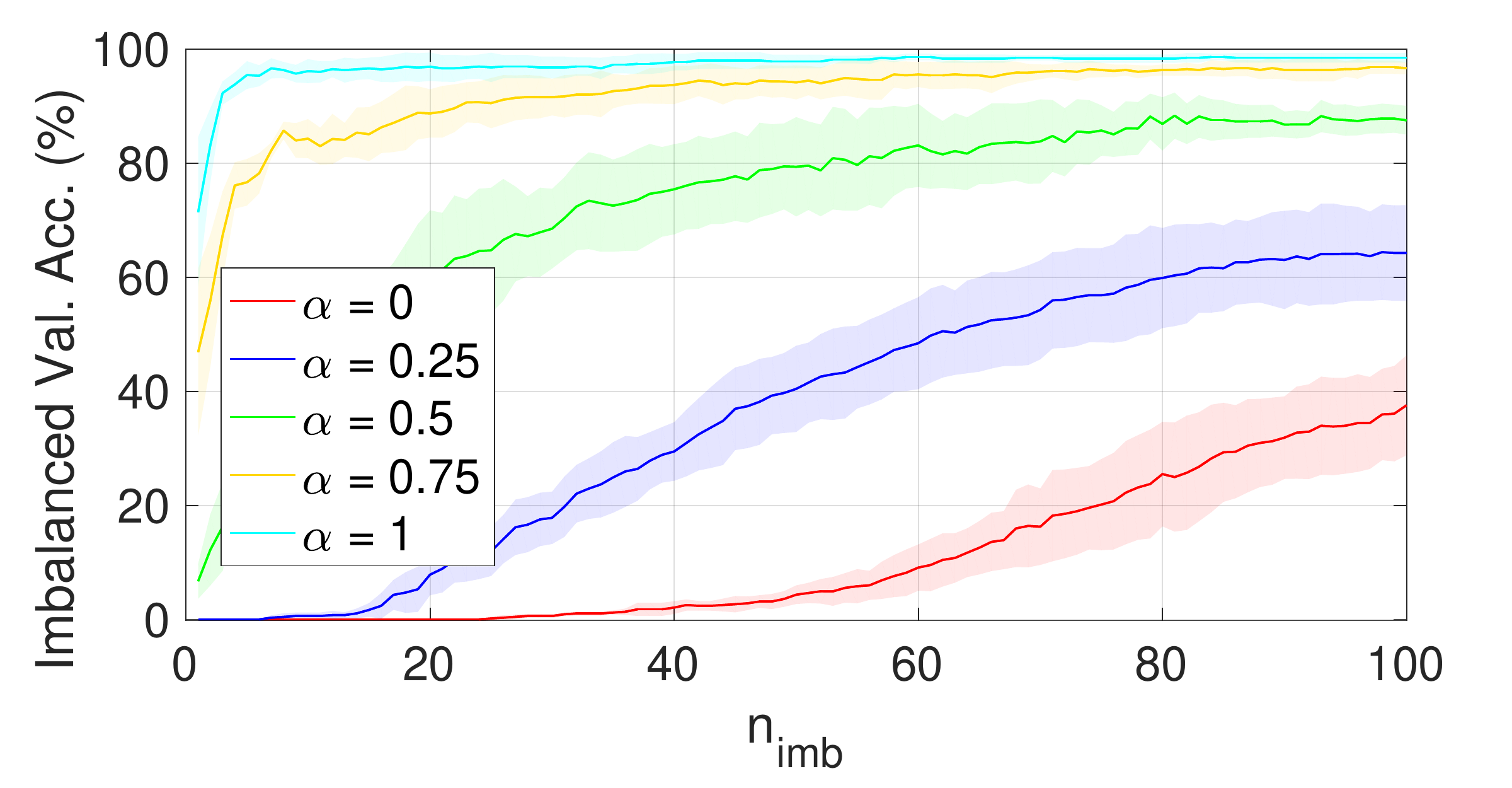}\\
\includegraphics[width = 0.4\textwidth]{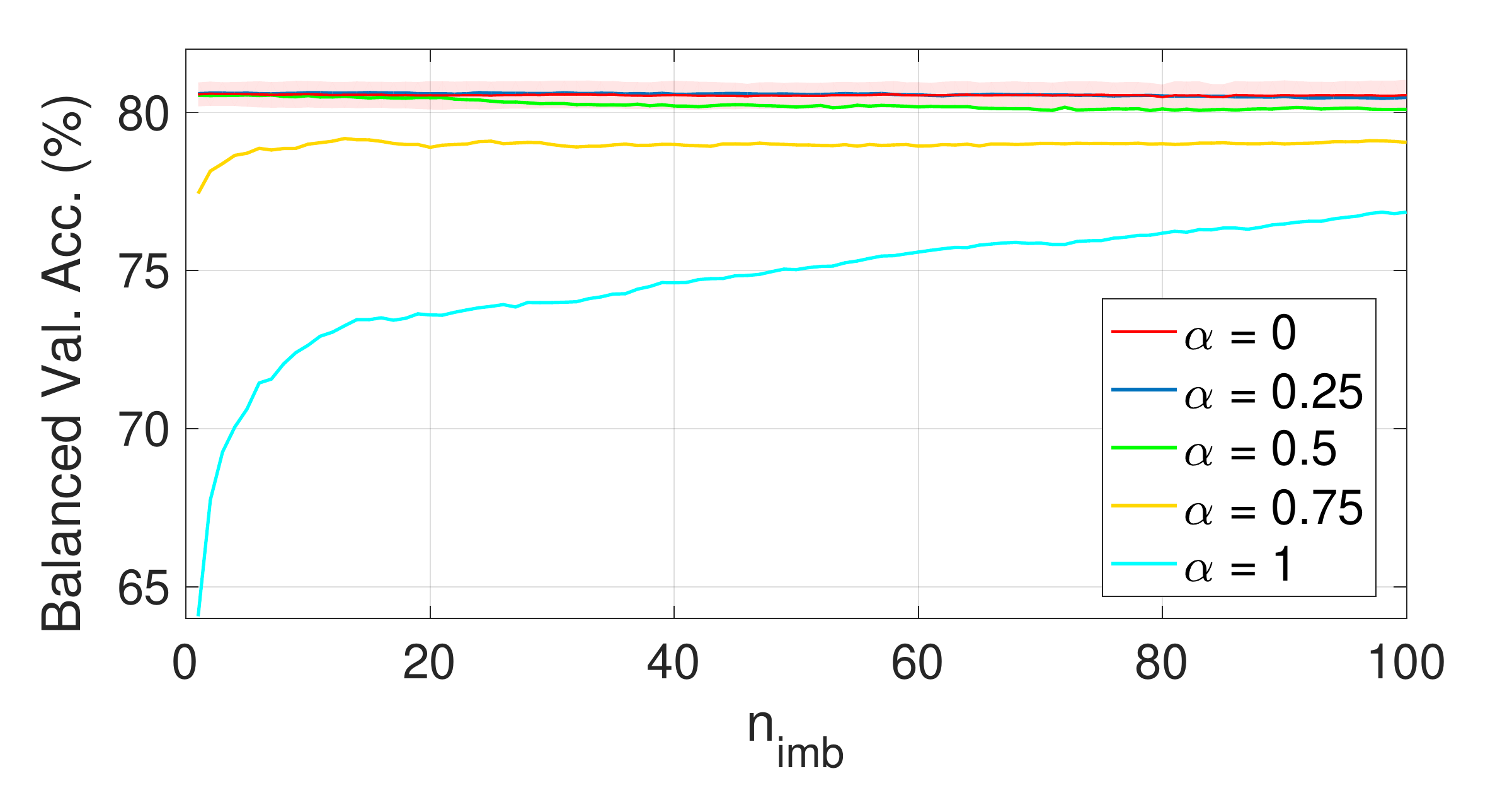}
\caption{Classification accuracy on \textit{iCubWorld28} imbalanced (Top) and balanced (Bottom) test classes for models trained according to Alg.\ref{alg:incremental_learning} with varying $\alpha$ and best $\lambda$ within a pre-defined range (chosen at each iteration and for each $\alpha$). Growing $\alpha$ from $0$ to $1$ allows to find a model that maintains the same performance on known classes while improving on the under-represented one.}
\label{fig:acc_modsel}
\end{figure}

Our strategy evaluates the accuracy of the candidate models on the incremental validation set, but {\it only for classes that have a sufficient number of examples} (e.g., classes with fewer examples than a pre-defined threshold are not used for validation). Then, we choose the model with largest $\alpha \in [0,1]$ for which such accuracy is higher or equal to the one measured for $\alpha = 0$, namely without coding. Indeed, as can be seen in Fig.~\ref{fig:acc_modsel} for validation experiments on {\it iCubWorld28}, as $\alpha$ grows from $0$ to $1$, the classification accuracy on the under-represented class increases, Fig.~\ref{fig:acc_modsel} (Top), while it decreases on the remaining ones, Fig.~\ref{fig:acc_modsel} (Bottom). Our heuristic chooses the best trade-off for $\alpha$ such that performance does not degrade on well-known classes, but at the same time it will often improve on the under-represented one.

\subsection{Results}

In Table~\ref{tab:accuracy} we report the results of the three methods on {\it MNIST}, {\it iCubWorld28} and {\it RGB-D} for a single under-represented class (digit ``8'', class $28$ and {\it tomato}, respectively). We observed a similar behaviour for other classes. We show both the accuracy on all classes (Total Acc., Left) and on the under-represented one (Imbalanced Acc., Right).
We note that, on the under-represented class, Alg.~\ref{alg:incremental_learning} ({\it RC}) consistently outperforms the RLSC baseline ({\it N}), which does not account for class imbalance and learns models that ignore the class. Also the total accuracy of {\it RC} results higher.  
Interestingly, on the two robotics tasks, {\it RC} outperforms the loss rebalancing approach ({\it RB}), particularly when very few examples of the under-represented class are available. This is favorable since, as we said, the rebalancing approach cannot be implemented incrementally (Sec.~\ref{sec:incremental_recoding}).


\begin{table}[t]                                     
\vspace*{2mm}
\centering        
\caption{Incremental classification accuracy for Na\"ive (N) RLSC, Rebalanced (RB) and Recoding (RC) (see Alg.~\ref{alg:incremental_learning}). Following the procedure described in Sec.~\ref{sec:model_selection}, we set $\alpha=0.7$ for {\it MNIST},  $\alpha=0.6$ for {\it iCubWorld28} and  $\alpha=0.7$ for {\it RGB-D}.}                                
\begin{adjustbox}{width=\linewidth,center}
\begin{tabular}{c c c c >{\columncolor{Gray}}c c c >{\columncolor{Gray}}c }                              
\toprule
\multirow{2}{*}{\textbf{Dataset}} & \multirow{2}{*}{$n_{imb}$} & \multicolumn{3}{c}{\textbf{Total Acc. (\%)}}  & \multicolumn{3}{c}{\textbf{Imbalanced Acc. (\%)}} \\ \cline{3-8}  
 \\ [-0.9em]
 & &  \textbf{N}  & \textbf{RB} & \textbf{RC}&  \textbf{N}  & \textbf{RB} & \textbf{RC} \\ 
\hline
\multirow{6}{*}{\shortstack{\textit{\textbf{MNIST}} \\ \\$n_{bal}$ \\ $=$ \\ $1000$ }}%
  & 1 & 79.2 $\pm$ 0.3 & \textbf{79.7 $\pm$ 0.4} & \textbf{79.7 $\pm$ 0.6} &     0.0 $\pm$     0.0 &  7.4 $\pm$ 7.7 & \textbf{9.5 $\pm$ 4.9} \\ 
  & 5 & 79.1 $\pm$ 0.3 & \textbf{82.5 $\pm$ 0.7} & 80.3 $\pm$ 0.6 &     0.0 $\pm$     0.0 & \textbf{39.6 $\pm$ 6.2 }& 17.5 $\pm$ 6.6 \\ 
  & 10 & 79.2 $\pm$ 0.3 & \textbf{83.6 $\pm$ 0.7} & 81.0 $\pm$ 0.6 &     0.0 $\pm$     0.0 & \textbf{49.5 $\pm$ 5.7} & 25.1 $\pm$ 5.3 \\ 
  & 50 & 79.2 $\pm$ 0.3 &  \textbf{85.5 $\pm$ 0.3} & 83.9 $\pm$ 0.5 &     0.0 $\pm$     0.0 & \textbf{73.5 $\pm$ 3.3} & 49.1 $\pm$ 3.5 \\ 
 & 100 & 79.2 $\pm$ 0.4 & \textbf{85.9 $\pm$ 0.4} & 85.1 $\pm$ 0.5 & 2.0 $\pm$ 0.9 & \textbf{75.5 $\pm$ 2.7} & 62.7 $\pm$ 2.9 \\ 
 & 500 & 85.5 $\pm$ 0.3 &\textbf{ 86.2 $\pm$ 0.3 }& \textbf{86.1 $\pm$ 0.3} & 66.9 $\pm$ 1.1 & \textbf{78.5 $\pm$ 0.9} & \textbf{77.8 $\pm$ 1.1} \\ 
\hline
\multirow{6}{*}{\shortstack{\textit{\textbf{iCub}} \\ \\$n_{bal}$ \\$=$ \\ $700$}}%
  &  1  & 77.6 $\pm$ 0.3 & 76.8 $\pm$ 0.1 & \textbf{77.7 $\pm$ 0.3} &     0.0 $\pm$     0.0 & 0.4 $\pm$ 0.6 &  \textbf{8.0 $\pm$ 11.4} \\ 
  & 5 & 77.6 $\pm$ 0.3 & 77.9 $\pm$ 0.1 & \textbf{78.6 $\pm$ 0.3} &     0.0 $\pm$     0.0 & 8.1 $\pm$ 3.9  & \textbf{38.5 $\pm$ 9.7} \\ 
  & 10 & 77.6 $\pm$ 0.3 & 78.3 $\pm$ 0.4 & \textbf{78.9 $\pm$ 0.2} &     0.0 $\pm$     0.0 & 23.7 $\pm$ 10.8 & \textbf{49.6 $\pm$ 5.6} \\ 
  & 50 & 77.7 $\pm$ 0.2 &   \textbf{80.0 $\pm$ 0.2} & \textbf{80.0 $\pm$ 0.1} & 5.4 $\pm$ 4.1 & 73.9 $\pm$ 7.3 &  \textbf{75.0 $\pm$ 5.5} \\ 
 & 100 & 78.6 $\pm$ 0.1 & \textbf{80.2 $\pm$ 0.1} & 80.1 $\pm$ 0.2 & 39.1 $\pm$ 3.6 & 85.9 $\pm$  4.0 & \textbf{86.5 $\pm$  3.0} \\ 
 & 500 & 80.2 $\pm$ 0.2 & \textbf{80.1 $\pm$ 0.1} & \textbf{80.1 $\pm$ 0.2} & 89.3 $\pm$ 2.5 & 93.8 $\pm$  2.0 & \textbf{94.8 $\pm$ 1.9} \\ 
\hline
\multirow{6}{*}{\shortstack{\textit{\textbf{RGB-D}} \\ \\$n_{bal} $\\ $=$ \\ $500$}} %
  & 1 & 80.4 $\pm$ 2.2 & 78.6 $\pm$ 3.2 & \textbf{83.3 $\pm$ 3.2} &     0.0 $\pm$     0.0 &  62.0 $\pm$ 42.1 & \textbf{72.2 $\pm$ 26.3} \\ 
  & 5 & 80.4 $\pm$ 2.2 &  83.0 $\pm$ 2.1 & \textbf{83.9 $\pm$ 2.6} &     0.0 $\pm$     0.0 & 91.7 $\pm$ 12.8 & \textbf{99.9 $\pm$ 0.3} \\ 
  & 10 & 80.4 $\pm$ 2.2 & \textbf{83.8 $\pm$ 1.8} & \textbf{83.6 $\pm$ 2.6} & 2.8 $\pm$ 2.4 & 94.7 $\pm$ 8.4 &     \textbf{100.0 $\pm$     0.0} \\ 
  & 50 & 82.3 $\pm$ 2.2 & \textbf{84.3 $\pm$ 1.9} & 83.5 $\pm$ 2.9 & 96.6 $\pm$ 3.7  &     \textbf{100.0$\pm$     0.0}  &     \textbf{100.0 $\pm$     0.0} \\ 
 & 100 & 82.4 $\pm$ 2.1 & \textbf{84.4 $\pm$ 2.0} & 83.5 $\pm$ 2.8 &     100.0 $\pm$     0.0 &    \textbf{ 100.0 $\pm$     0.0 } &     \textbf{100.0 $\pm$     0.0} \\ 
 & 500 & 82.3 $\pm$ 2.1 & \textbf{84.1 $\pm$ 2.0} & \textbf{84.1 $\pm$ 2.8}  &     100.0 $\pm$     0.0 &     \textbf{100.0 $\pm$     0.0}  &     \textbf{100.0 $\pm$     0.0} \\ 
\bottomrule
\end{tabular}                                     
\end{adjustbox}
\label{tab:accuracy}                        
\end{table}

\begin{figure}[t]
\vspace{1\baselineskip}
\centering
\includegraphics[width = 0.165\textwidth]{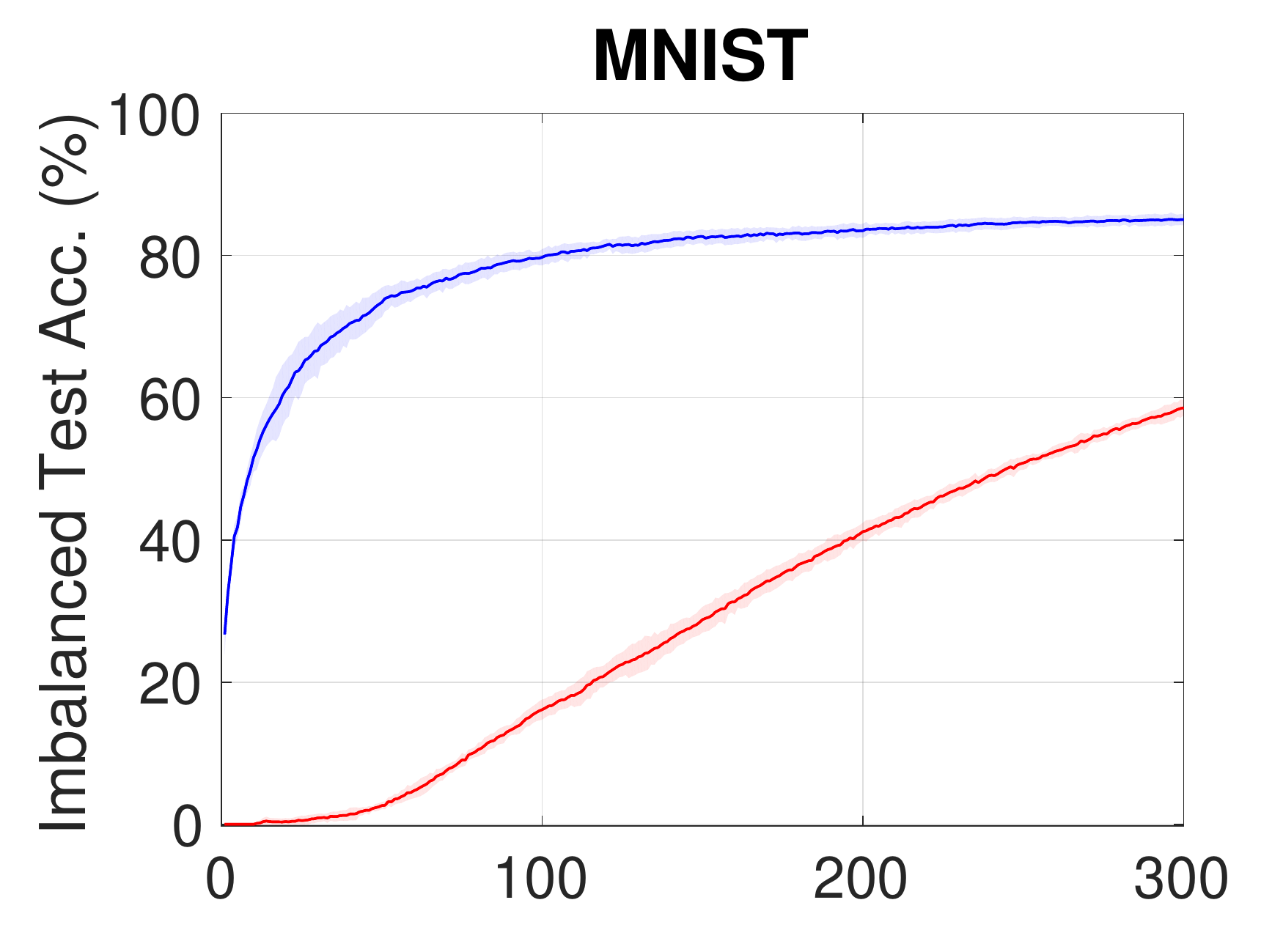}%
\includegraphics[width = 0.165\textwidth]{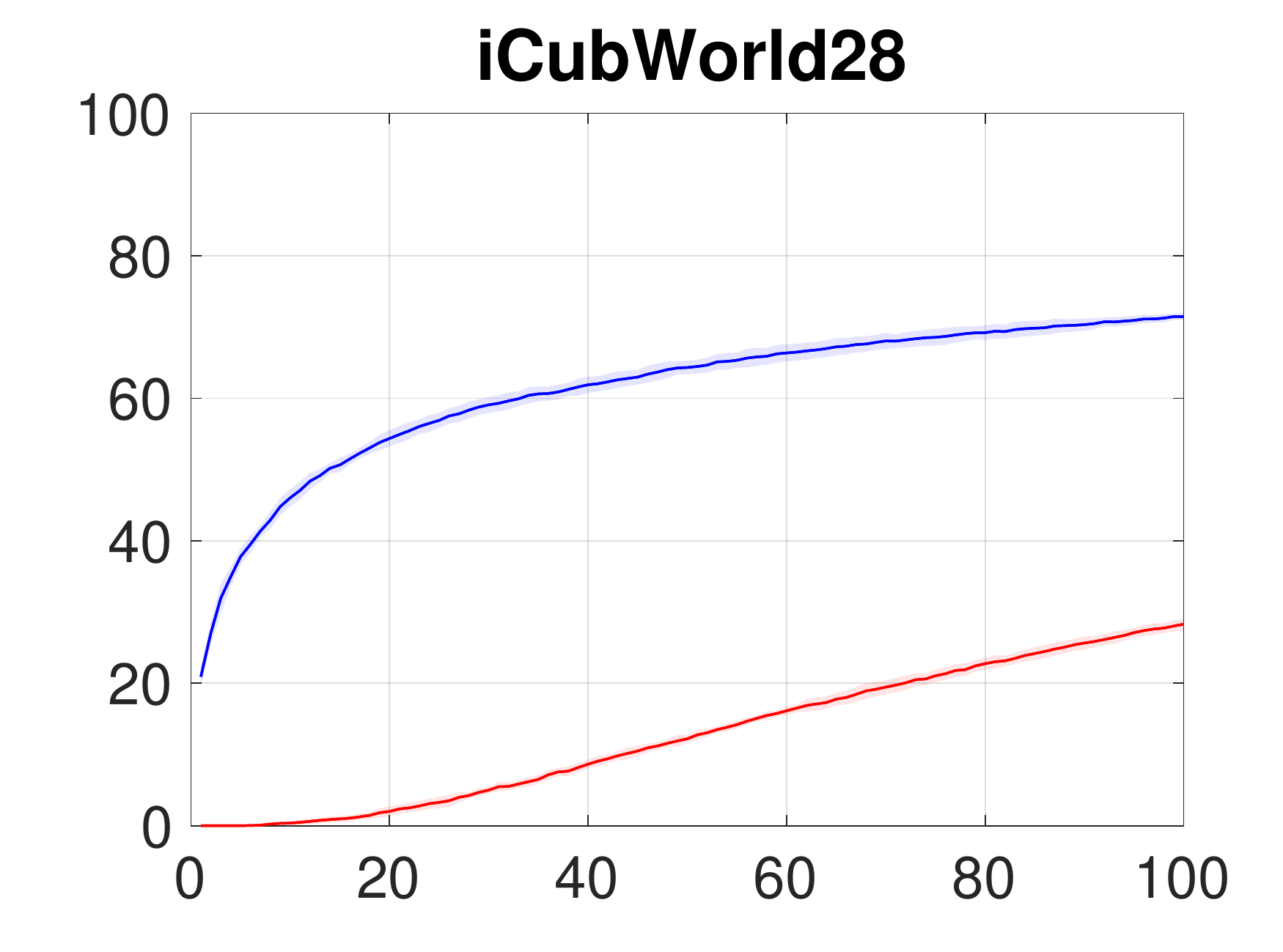}%
\includegraphics[width = 0.165\textwidth]{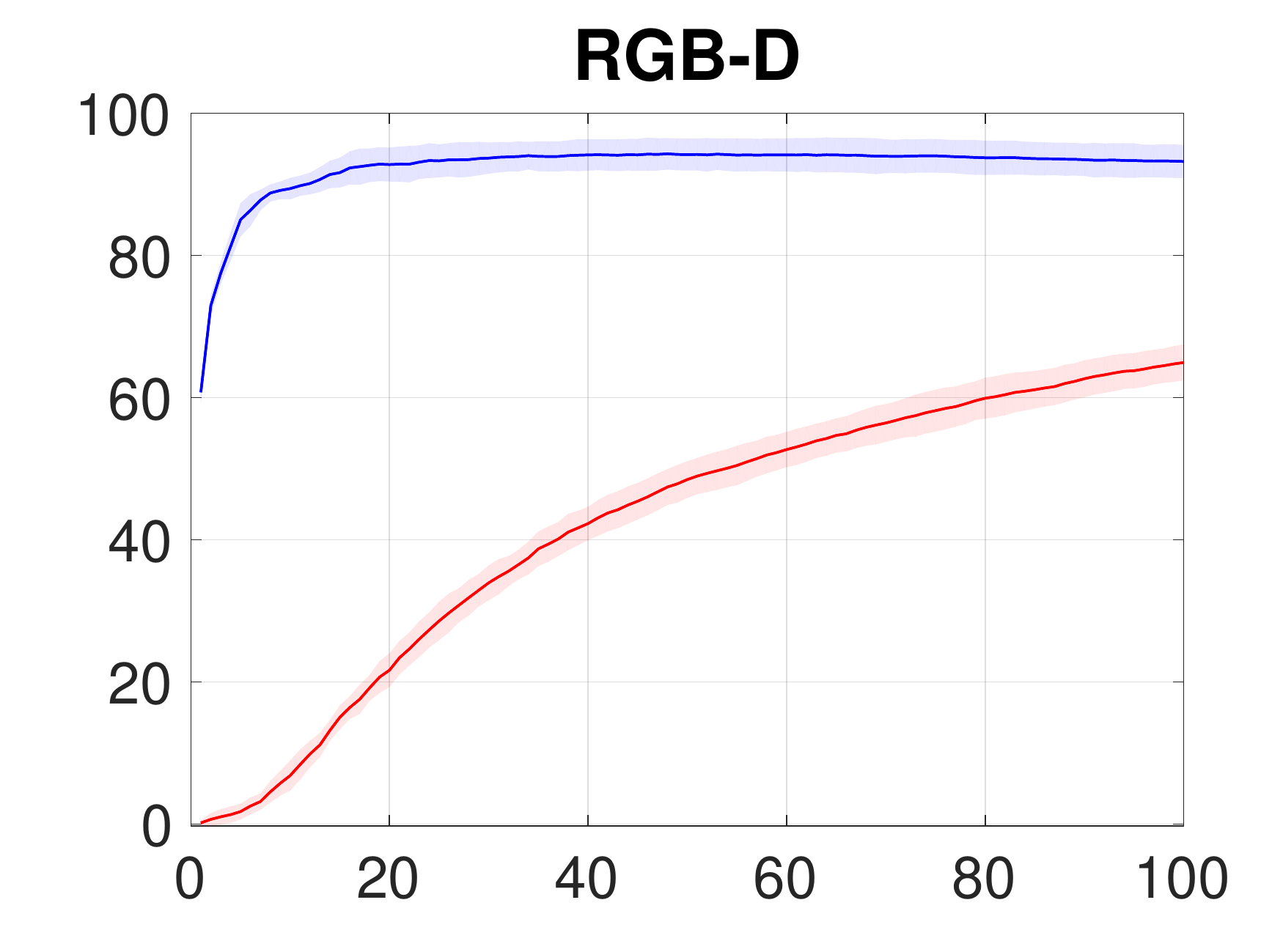}\\
\includegraphics[width = 0.165\textwidth]{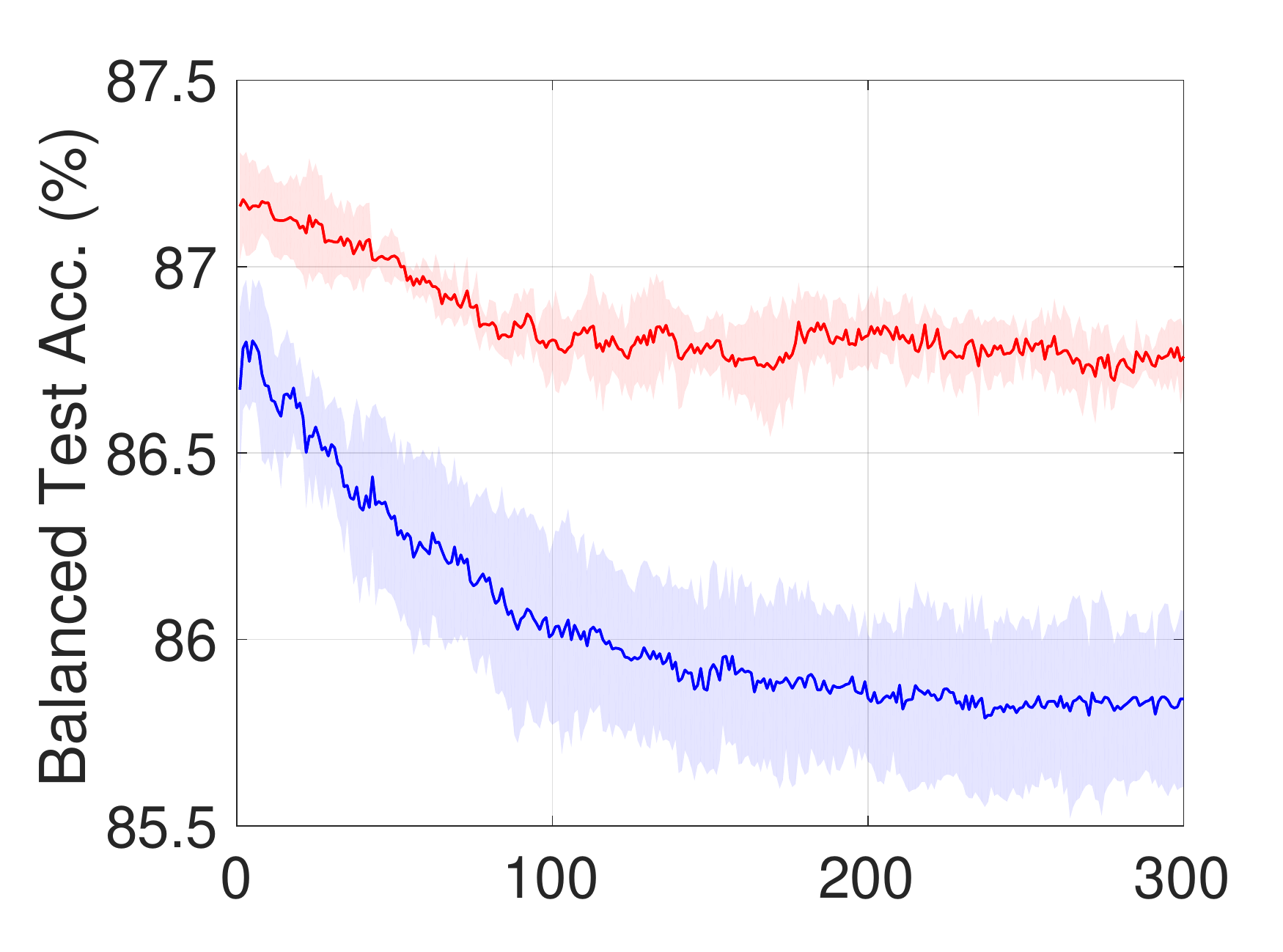}%
\includegraphics[width = 0.165\textwidth]{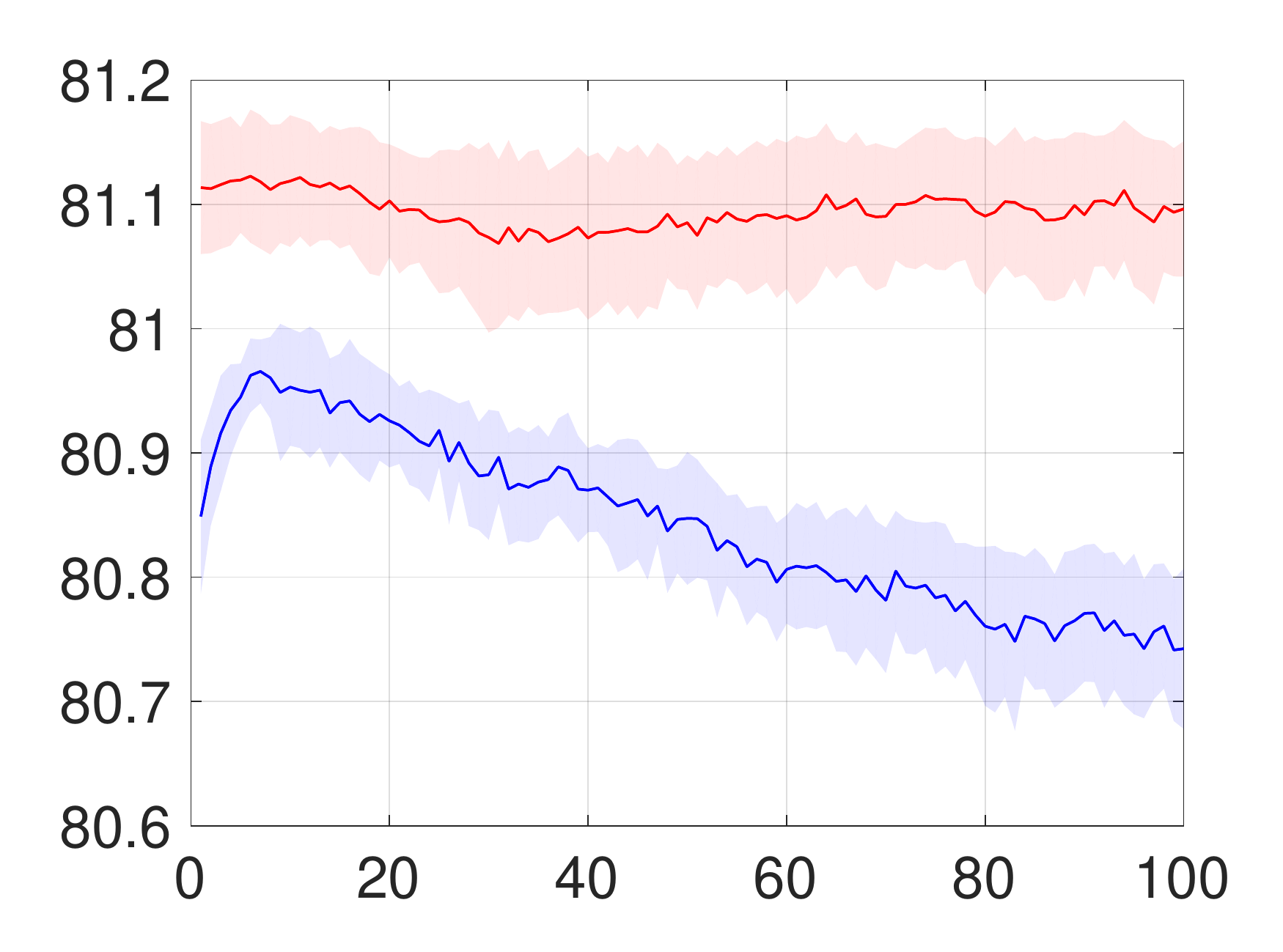}%
\includegraphics[width = 0.165\textwidth]{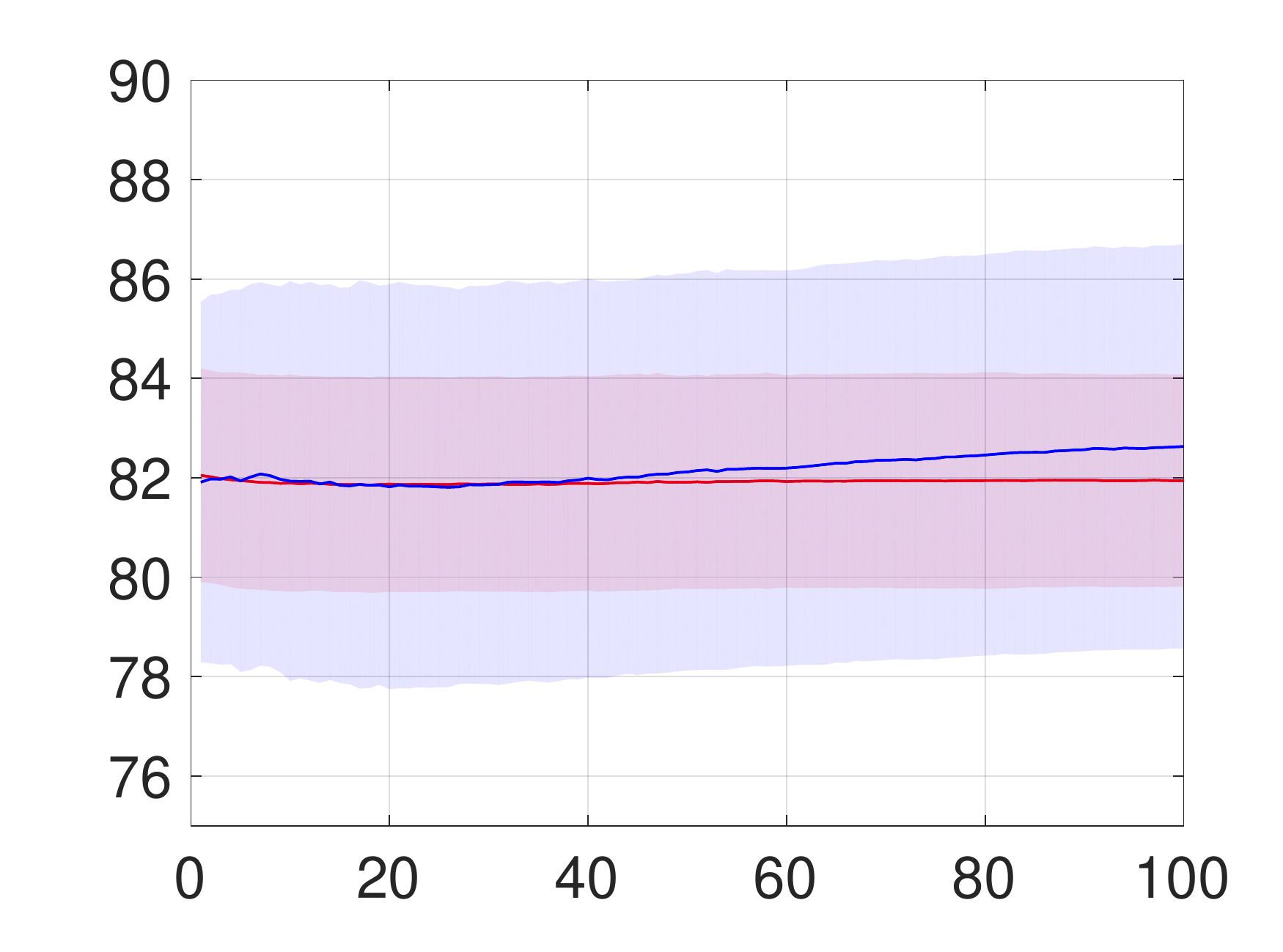}\\
\includegraphics[width = 0.165\textwidth]{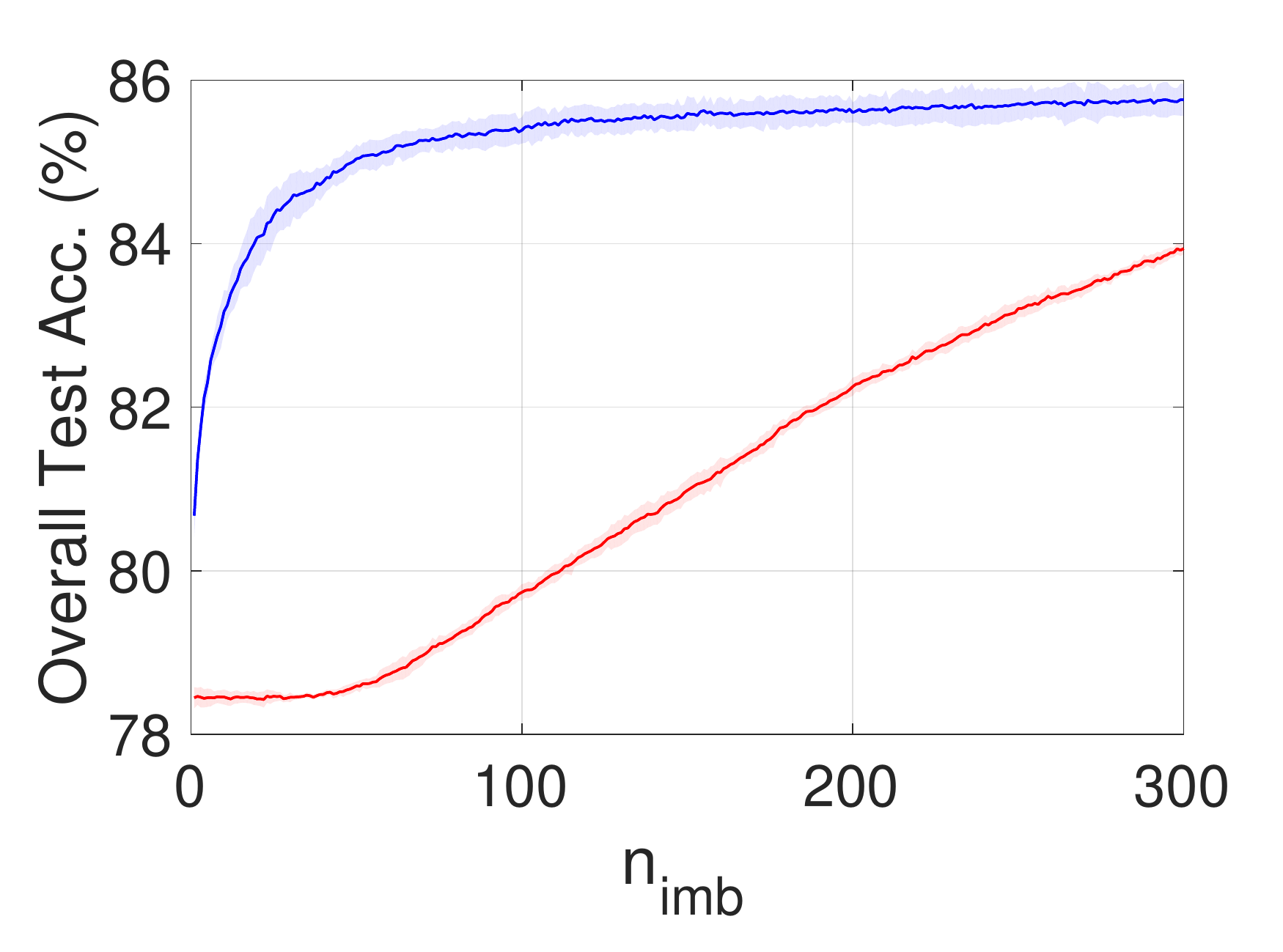}%
\includegraphics[width = 0.165\textwidth]{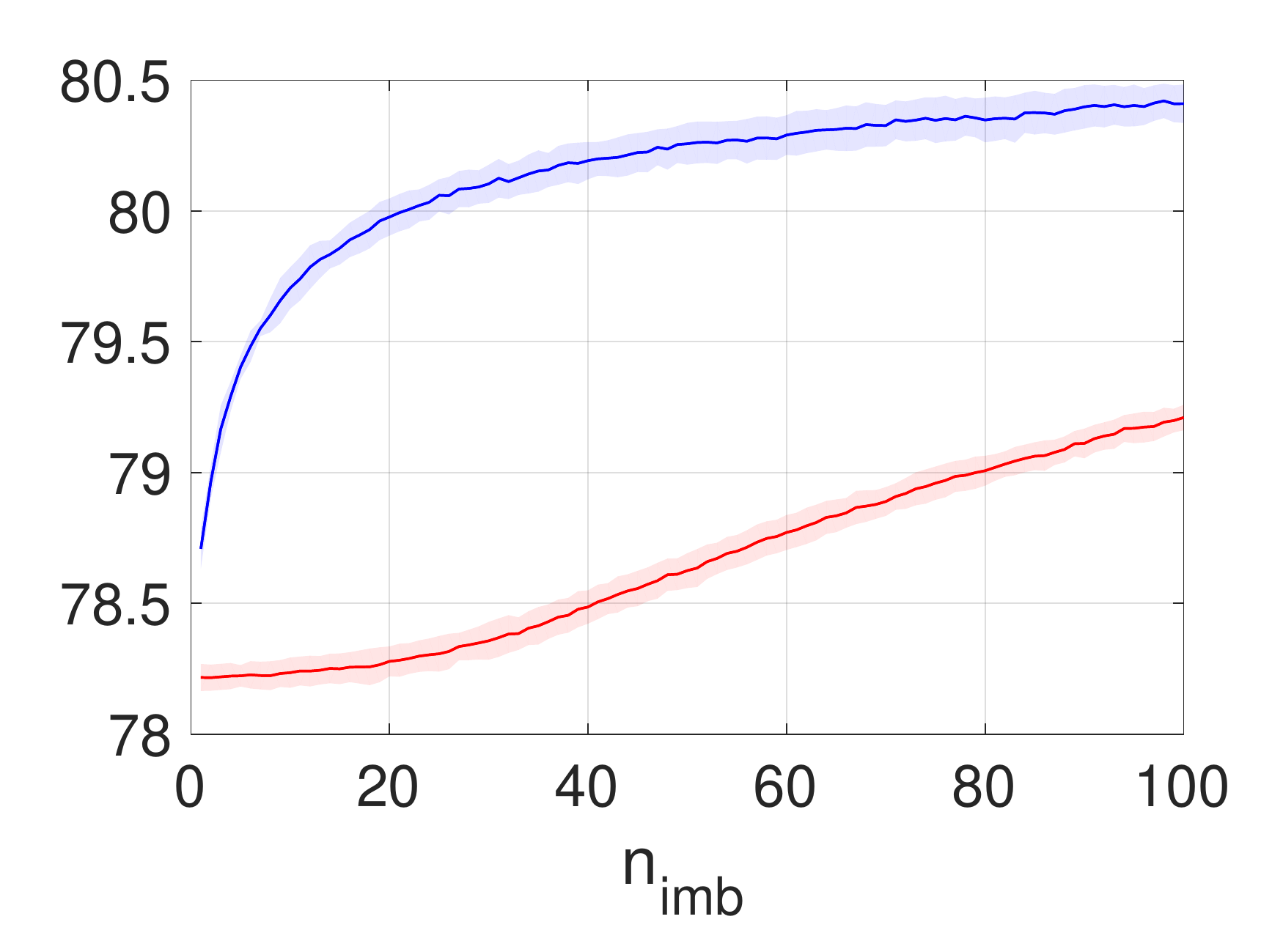}%
\includegraphics[width = 0.165\textwidth]{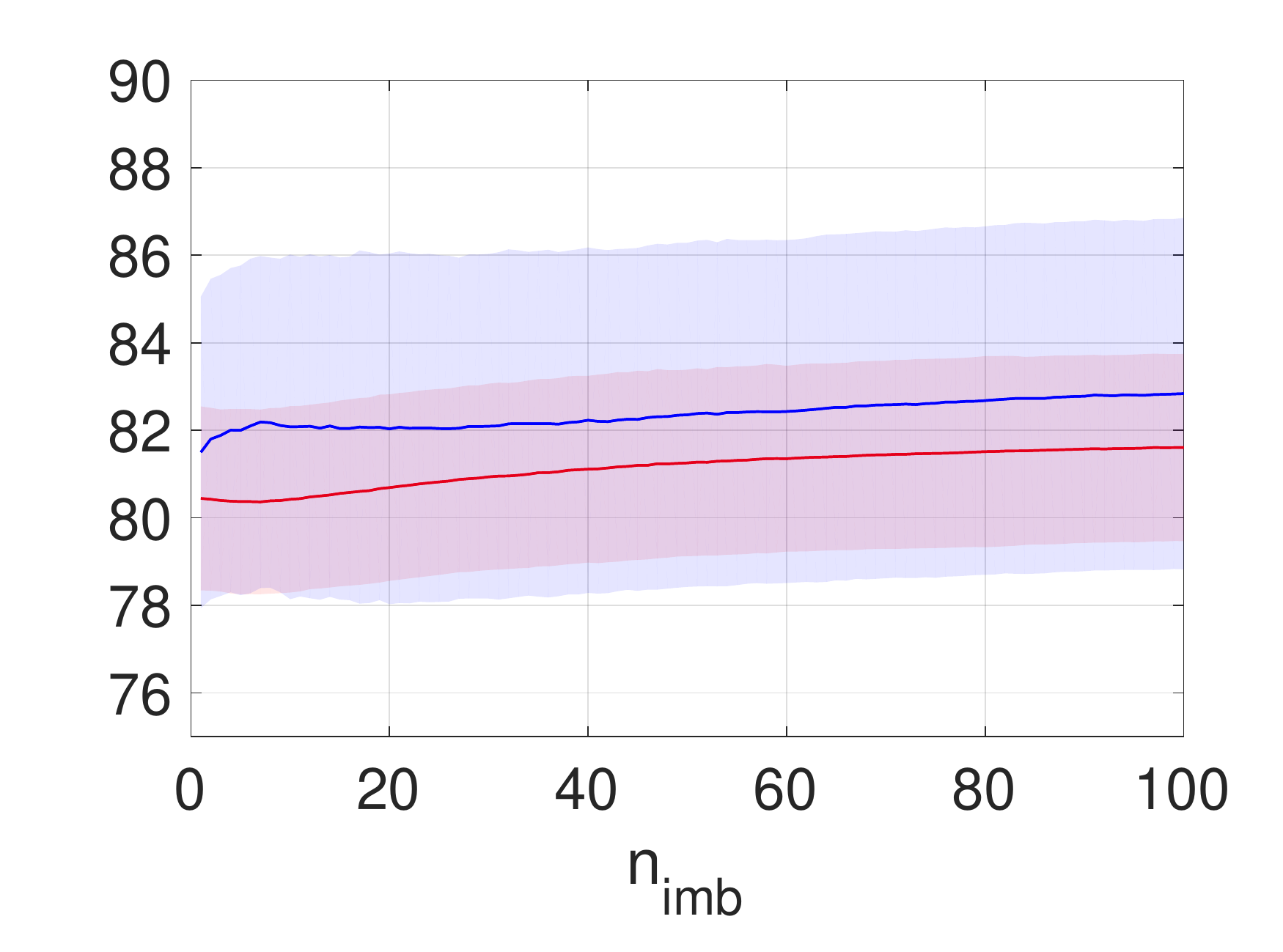}\\
\caption{Average test classification accuracy of the standard incremental RLSC (Red) and the variant proposed in this work (Blue) over the imbalanced (Top), balanced (Middle) and all (Bottom) classes. The models are incrementally trained as $n_{imb}$ grows, as described in Sec.~\ref{sec:protocol}.}
\label{fig:acc}
\end{figure}

To offer a clear intuition of the improvement provided by our method, in Fig.~\ref{fig:acc} we show the accuracy of Na\"ive RLSC (Red) and Alg.~\ref{alg:incremental_learning} (Blue), separately on the under-represented class (Top), the balanced classes (Middle), and all classes (Bottom), as they are trained on new examples. For this experiment we let each class be under-represented and averaged the results.
It can be noticed that Alg.~\ref{alg:incremental_learning} is much better on the imbalanced class, while being comparable on the balanced ones, resulting in overall improved performance.

We point out that the total accuracy on all datasets for $n_{imb}=500$ is comparable with the state of the art. Indeed, on {\it MNIST} we achieve $\sim 86 \%$ accuracy, which is slightly lower than the one reported in~\cite{lecun1998gradient} for a linear classifier on top of raw pixels (this is reasonable, since we are using much fewer training examples).
The total accuracy of Alg.~\ref{alg:incremental_learning} on \textit{RGB-D} is approximately $84\%$, which is comparable with the state of the art on this dataset~\cite{schwarz2015}. 
On the {\it iCubWorld28} dataset we achieve $\sim 80\%$ accuracy, which is in line with the results reported in Fig. 8 of~\cite{pasquale2015b} (extended version of~\cite{pasquale2015teaching}).


\section{Conclusion}
\label{sec:conclusions}

In this paper we addressed the problem of learning online with an increasing number of classes. Motivated by the visual recognition scenario in lifelong robot learning, we focused on issues related to class imbalance, which naturally arises when a new object/category is observed for the first time. To address this problem, we proposed a variant of the recursive Regularized Least Squares for Classification (RLSC) algorithm that (i) incorporates new classes incrementally and (ii) dynamically applies class recoding when new examples are observed. Updates are performed in constant time with respect to the growing number of training examples. We evaluated the proposed algorithm on a standard machine learning benchmark and on two datasets for visual recognition in robotics, showing that our approach is indeed favorable in online settings when classes are imbalanced.

We note that, in principle, for the experiments where we used features extracted from a Convolutional Neural Network, we could have also directly trained the network online, by Stochastic Gradient Descent (backpropagation). 
While works empirically investigating this end-to-end approach in settings where new classes are to be progressively included into the model exist~\cite{kading2016fine}, this is still a largely unexplored field, the study of which is not in the scope of this work. The method we propose allows to update a predictor without using training data from previous classes in a fast and stable way, and, by relying on rich deep representations learned offline, is proven to be competitive with the state of the art, while being more suitable for online applications.

Future research will focus on strategies to exploit knowledge of known classes to improve classification accuracy on new ones, following recent work~\cite{tommasi2010,tommasi2012,kuzborskij2013,sunneol2016}. 

 \section*{Acknowledgment}
The work described in this paper is supported by the Center for Brains, Minds and Machines, funded
by NSF STC award CCF-1231216 and by FIRB project RBFR12M3AC, funded by the Italian Ministry of Education,
University and Research. We acknowledge NVIDIA Corporation for the donation of the Tesla k40 GPU
used for this research.


\bibliographystyle{IEEEtran}
\bibliography{iros2016}

\end{document}